\documentclass{article}
\usepackage{times}
\usepackage{fullpage}
\usepackage{amsfonts}
\usepackage{amssymb}
\usepackage{amsmath}
\usepackage{epsfig}
\usepackage{color}
\usepackage{array}
\usepackage{pseudocode}
\def\jump{\vskip0.05in}

\def\wmu{{\widehat \mu}}

\def\qed{\vrule height8pt width3pt depth0pt}

\def\P{{\mathbb P}}

\def\E{{\mathbb E}}
\def\R{{\mathbb R}}

\def\wX{{\widetilde{X}}} 
\def\wS{{\widetilde{S}}} 
 
\def\wx{{\widetilde{x}}}

\def\wmu{{\widetilde{\mu}}}
\def\med{{\mbox{\rm median}}}
\def\mean{{\mbox{\rm mean}}}

\def\cov{{\mbox{\rm cov}}}
\def\var{{\mbox{\rm var}}}

\newtheorem{thm}{Theorem}
\newtheorem{lemma}[thm]{Lemma}
\newtheorem{cor}[thm]{Corollary}

\newtheorem{defn}[thm]{Definition}

\newenvironment{proof}{\noindent{\it Proof.} }{\qed\jump}

\begin{document}

\title{Random projection trees for vector quantization}

\author{Sanjoy~Dasgupta~and~Yoav~Freund \thanks{Both authors are with the Department of Computer Science and Engineering, University of California, San Diego. Email: {\tt dasgupta,yfreund@cs.ucsd.edu}.}}

\maketitle

\begin{abstract}
A simple and computationally efficient scheme for tree-structured vector 
quantization is presented. Unlike previous methods, its quantization error 
depends only on the intrinsic dimension of the data distribution, rather 
than the apparent dimension of the space in which the data happen to lie.
\end{abstract}

\section{Introduction}

For a distribution $P$ on $\R^D$, the {\it $k$th quantization error} is 
commonly defined as
$$ \inf_{\mu_1, \ldots, \mu_k \in \R^D} 
\E \left[ \min_{1 \leq j \leq k} \|X - \mu_j\|^2 \right],$$
where $\| \cdot \|$ denotes Euclidean norm and the expectation is over 
$X$ drawn at random from $P$. It is known \cite{P82} that this
infimum is realized, though perhaps not uniquely, by some set of 
points $\mu_1, \ldots, \mu_k$, called a {\it $k$-optimal set of 
centers}. The resulting quantization error has been shown to be roughly 
$k^{-2/D}$ under a variety of different assumptions on $P$ \cite{GL00}. 
This is discouraging when $D$ is high. For instance, 
if $D = 1000$, it means that to merely 
halve the error, you need $2^{500}$ times as many codewords! In short, 
vector quantization is susceptible to the same {\it curse of dimensionality} 
that has been the bane of other nonparametric statistical methods.

A recent positive development in statistics and machine learning has 
been the realization that a lot of data that superficially lie in a 
high-dimensional space $\R^D$, actually have low {\it intrinsic} 
dimension, in the sense of lying close to a manifold of dimension 
$d \ll D$. We will give several examples of this below. There has 
thus been a huge interest in algorithms that {\it learn} this manifold 
from data, with the intention that future data can then be transformed 
into this low-dimensional space, in which the usual nonparametric 
(and other) methods will work well \cite{TdSL00,RS00,BN03}.

\begin{figure}
\begin{center}
\includegraphics[height=5cm]{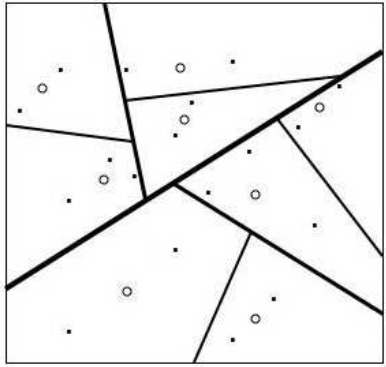}
\end{center}
\caption{Spatial partitioning of $\R^2$ induced by an RP tree with three levels.
The dots are data points; each circle represents the mean of the vectors in one 
cell.}
\label{fig:rppic}
\end{figure}

In this paper, we are interested in techniques that automatically adapt
to intrinsic low dimensional structure without having to explicitly learn
this structure. We describe a tree-structured vector quantizer whose 
quantization error is $k^{-1/O(d)}$; that is to say, 
its error rate depends only on the low intrinsic dimension rather than 
the high apparent dimension. The quantizer is based on a hierarchical 
decomposition of $\R^D$: first the entire space is split into two pieces, 
then each of these pieces is further split in two, and so on, until a 
partition of $k$ cells is reached. Each codeword is the mean of the 
distribution restricted to one of these cells.

Tree-structured vector quantizers abound; the power of our approach comes
from the particular splitting method. To divide a region $S$ into two, 
we pick a random direction from the surface of the unit sphere in $\R^D$, 
and split $S$ at the median of its projection onto this direction 
(Figure~\ref{fig:rppic}). We call the resulting spatial partition a 
{\it random projection tree} or {\it RP tree}.

At first glance, it might seem that a better way to split a region is to 
find the 2-optimal set of centers for it. However, as we explain below, 
this is an NP-hard optimization problem, and is therefore unlikely to be 
computationally tractable. Although there are several algorithms that 
attempt to solve this problem, such as Lloyd's method \cite{MacQ67,L82}, 
they are not in 
general able to find the optimal solution. In fact, they are often 
far from optimal.

For our random projection trees, we show that if the data have intrinsic 
dimension $d$ (in a sense we make precise below), then each split 
pares off about a $1/d$ fraction of the quantization error. Thus, after 
$\log k$ levels of splitting, there are $k$ cells and the quantization 
error is of the form $(1-1/d)^{\log k} = k^{-1/O(d)}$. There is no 
dependence at all on the extrinsic dimensionality $D$.

\section{Detailed overview}

\subsection{Low-dimensional manifolds}

The increasing ubiquity of massive, high-dimensional data sets has focused 
the attention of the statistics and machine learning communities on the curse 
of dimensionality. A large part of this effort is based on exploiting the 
observation that many high-dimensional data sets have low {\it intrinsic 
dimension}. This is a loosely defined notion, which is typically used to mean 
that the data lie near a smooth low-dimensional manifold. 

For instance, suppose that you wish to create realistic animations by collecting
human motion data and then fitting models to it. A common method for collecting 
motion data is to have a person wear a skin-tight suit with high contrast 
reference points printed on it. Video cameras are used to track the 3D 
trajectories of the reference points as the person is walking or running.  
In order to ensure good coverage, a typical suit has about $N = 100$ reference 
points. The position and posture of the body at a particular point of time is 
represented by a $(3N)$-dimensional vector. However, despite this seeming high
dimensionality, the number of degrees of freedom is small, corresponding to 
the dozen-or-so joint angles in the body.  The positions of the reference 
points are more or less deterministic functions of these joint angles. 

Interestingly, in this example the intrinsic dimension becomes even smaller 
if we {\em double} the dimension of the embedding space by including for 
each sensor its relative velocity vector. In this space of dimension $6N$ 
the measured points will lie very close to the {\em one} dimensional manifold 
describing the combinations of locations and speeds that the limbs go through 
during walking or running.

To take another example, a speech signal is commonly represented by a 
high-dimensional time series: the signal is broken into overlapping windows,
and a variety of filters are applied within each window. Even richer 
representations can be obtained by using more filters, or by concatenating 
vectors corresponding to consecutive windows. Through all this, the intrinsic
dimensionality remains small, because the system can be described
by a few physical parameters describing the configuration of the speaker's
vocal apparatus.

In machine learning and statistics, almost all the work on exploiting intrinsic 
low dimensionality consists of algorithms for learning the structure of these
manifolds; or more precisely, for learning embeddings of these manifolds into
low-dimensional Euclidean space. Our contribution is a simple and compact
data structure that automatically exploits the low intrinsic dimensionality 
of data on a local level without having to explicitly learn the global manifold 
structure.

\subsection{Defining intrinsic dimensionality}

\begin{figure}
\begin{center}
\includegraphics[height=4.5cm]{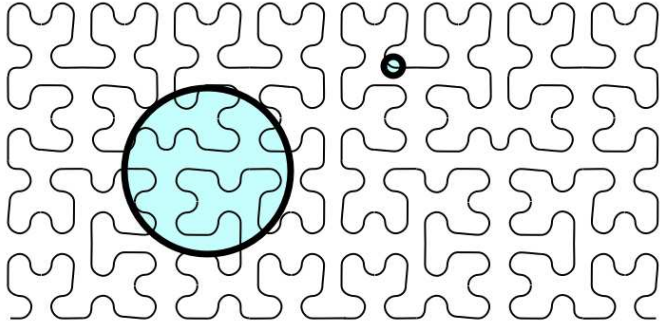}
\end{center}
\caption{Hilbert's space filling curve. Large neighborhoods look $2$-dimensional,
smaller neighborhoods look $1$-dimensional, and even smaller neighborhoods would
consist mostly of measurement noise and would therefore again be $2$-dimensional.}
\label{fig:curvature}
\end{figure}

Low-dimensional manifolds are our inspiration and source of intuition, but
when it comes to precisely defining intrinsic dimension for data analysis, 
the differential geometry concept of manifold is not entirely suitable. 
First of all, any data set lies on a one-dimensional manifold, as evidenced
by the construction of space-filling curves. Therefore, some bound on
curvature is implicitly needed. Second, and more important, it is unreasonable 
to expect data to lie {\it exactly} on a low-dimensional manifold. At a certain 
small resolution, measurement error and noise make any data set full-dimensional. 
The most we can hope is that the data distribution is concentrated {\it near} 
a low-dimensional manifold of bounded curvature (Figure~\ref{fig:curvature}).

We address these various concerns with a statistically-motivated notion 
of dimension: we say that a set $S$ has {\it local covariance dimension} 
$(d,\epsilon,r)$ if neighborhoods of radius $r$ have $(1-\epsilon)$ fraction 
of their variance concentrated in a $d$-dimensional subspace. To make this 
precise, start by letting $\sigma_1^2, \sigma_2^2, \ldots, \sigma_D^2$ denote 
the eigenvalues of the covariance matrix; these are the variances in each of 
the eigenvector directions.
\begin{defn}
Set $S \subset \R^D$ has {\it local covariance dimension}
$(d,\epsilon, r)$ if its restriction to any ball of radius $r$ has 
covariance matrix whose largest $d$ eigenvalues satisfy
$$ \sigma_1^2 + \cdots + \sigma_d^2 \ \geq \ 
(1 - \epsilon) \cdot (\sigma_1^2 + \cdots + \sigma_D^2) .$$
\end{defn}

\subsection{Random projection trees}

Our new data structure, the random projection tree, is built by recursive 
binary splits. The core tree-building algorithm is called {\sc MakeTree}, 
which takes as input a data set $S \subset \R^D$, and repeatedly calls 
a splitting subroutine {\sc ChooseRule}.

\medskip
\begin{pseudocode}[display]{}{}
\PROCEDURE{MakeTree}{S}
\IF |S| < MinSize \CTHEN \RETURN{Leaf} \\
Rule \GETS \CALL{ChooseRule}{S} \\
LeftTree \GETS \CALL{MakeTree}{\{x \in S: Rule(x) = \mbox{true}\}} \\
RightTree \GETS \CALL{MakeTree}{\{x \in S: Rule(x) = \mbox{false}\}} \\
\RETURN{[Rule, LeftTree, RightTree]}
\ENDPROCEDURE
\end{pseudocode}

The RP tree has two types of split. Typically, a direction 
is chosen uniformly at random from surface of the unit sphere and 
the cell is split at its median, by a hyperplane orthogonal to this 
direction. Occasionally, a different type of split is used, in which 
a cell is split into two pieces based on distance from the mean. 

\medskip
\begin{pseudocode}[display]{}{}
\PROCEDURE{ChooseRule}{S}
\IF \Delta^2(S) \leq c \cdot \Delta_A^2(S)
\THEN
\BEGIN
\mbox{choose a random unit direction $v$} \\
Rule(x) := x \cdot v \leq \mbox{median}(\{z \cdot v: z \in S\}) \\
\END
\ELSE
\BEGIN
Rule(x) := \\ \|x - \mean(S)\| \leq \med(\{\|z - \mean(S)\|: z \in S\})
\END \\
\RETURN{Rule}
\ENDPROCEDURE
\end{pseudocode}
In the code, $c$ is a constant, $\Delta(S)$ is the diameter of $S$ 
(the distance between the two furthest points in the set), and 
$\Delta_A(S)$ is the {\it average} diameter, that is, the average 
distance between points of $S$:
$$ \Delta_A^2(S) = \frac{1}{|S|^2} \sum_{x,y \in S} \|x-y\|^2 .$$

\subsection{Main result}

Suppose an RP tree is built from a data set $S \subset \R^D$, not necessarily
finite. If the tree has $k$ levels, then it partitions the space into $2^k$ 
cells. We define the {\it radius} of a cell $C \subset \R^D$ to be the smallest
$r > 0$ such that $S \cap C \subset B(x,r)$ for some $x \in C$.

Recall that an RP tree has two different types of split; let's
call them splits {\it by distance} and splits {\it by projection}.

\begin{thm}
There are constants $0 < c_1, c_2, c_3 < 1$ with the following property.
Suppose an RP tree is built using data set $S \subset \R^D$.
Consider any cell $C$ of radius $r$, such that $S \cap C$ has local covariance 
dimension $(d, \epsilon, r)$, where $\epsilon < c_1$. Pick a point 
$x \in S \cap C$ at random, and let $C'$ be the cell that contains it at 
the next level down.
\begin{itemize}
\item If $C$ is split by distance,
$ \E \left[ \Delta(S \cap C') \right] \leq c_2 \Delta(S \cap C).$
\item If $C$ is split by projection, then 
$\E \left[ \Delta_A^2(S \cap C') \right] 
\leq 
\left( 1 - (c_3/d)\right) \Delta_A^2(S \cap C).$
\end{itemize}
In both cases, the expectation is over the randomization in splitting $C$
and the choice of $x \in S \cap C$.
\label{thm:main2}
\end{thm}

\subsection{The hardness of finding optimal centers}

Given a data set, the optimization problem of finding a $k$-optimal set of 
centers is called the $k$-means problem. Here is the formal definition.
\begin{quote}
{\sc $k$-means clustering}

{\it Input:} Set of points $x_1, \ldots, x_n \in \R^D$; integer $k$.

{\it Output:} A partition of the points into clusters $C_1, \ldots, C_k$,
along with a center $\mu_j$ for each cluster, so as to minimize
$$ \sum_{j=1}^k \sum_{i \in C_j} \|x_i - \mu_j \|^2 .$$
\end{quote}
The typical method of approaching this task is to apply Lloyd's algorithm 
\cite{MacQ67,L82}, and usually this algorithm is itself called $k$-means. 
The distinction between the two is particularly important to make because
Lloyd's algorithm is a heuristic that often returns a suboptimal solution
to the $k$-means problem. Indeed, its solution is often very far from 
optimal.

What's worse, this suboptimality is not just a problem with Lloyd's algorithm,
but an inherent difficulty in the optimization task. {\sc $k$-means clustering} 
is an NP-hard optimization problem, which means that it is very unlikely that 
there exists an efficient 
algorithm for it. To explain this a bit more clearly, we delve briefly into 
the theory of computational complexity.

The running time of an algorithm is typically measured as a function of
its input/output size. In the case of $k$-means, for instance, it would be given
as a function of $n$, $k$, and $D$. An efficient algorithm is one whose running 
time scales {\it polynomially} with the problem size. For instance, there are 
algorithms for sorting $n$ numbers which take time proportional to $n \log n$; 
these qualify as efficient because $n\log n$ is bounded above by a polynomial 
in $n$.

For some optimization problems, the best algorithms we know 
take time {\it exponential} in problem size. The famous
traveling salesman problem (given distances between $n$ cities, plan a 
circular route through them so that each city is visited once and the
overall tour length is minimized) is one of these. There are 
various algorithms for it that take time proportional to $2^n$ (or worse): 
this means each additional city causes the running time to be doubled!
Even small graphs are therefore hard to solve.

This disturbing lack of an efficient algorithm is not limited to just 
a few pathological optimization tasks. Rather, it is an epidemic across
the entire spectrum of computational tasks, one that afflicts 
thousands of the problems we most urgently want to solve. Amazingly,
it has been shown that the fates of these diverse problems (called 
{\it NP-complete} problems) are linked: either {\it all} of them admit 
efficient algorithms, or none of them do! The mathematical community 
strongly believes the latter to be the case, although it is has not been 
proved. Resolving this question is one of the seven ``grand challenges'' 
of the new millenium identified by the Clay Institute.

In Appendix II, we show the following.
\begin{thm}
{\sc $k$-means clustering} is an NP-hard optimization problem, even
if $k$ is restricted to 2.
\label{thm:hardness}
\end{thm}
Thus we cannot expect to be able to find a $k$-optimal set of centers; the 
best we can hope is to find some set of centers that achieves roughly the 
optimal quantization error.

\subsection{Related work}

\subsubsection*{Quantization} 

The literature on vector quantization is substantial;
see the wonderful survey of Gray and Neuhoff \cite{GN98} for a comprehensive 
overview. In the most basic setup, there is some distribution $P$ over $\R^D$ 
from which random vectors are generated and observed, and the goal is 
to pick a finite codebook $C \subset \R^D$ and an encoding function 
$\alpha: \R^D \rightarrow C$ such that $x \approx \alpha(x)$ for typical vectors 
$x$. The quantization error is usually measured by squared loss, 
$\E \|X - \alpha(X)\|^2$. An obvious choice is to let $\alpha(x)$ be the nearest 
neighbor of $x$ in $C$. However, the number of codewords is often so enormous 
that this nearest neighbor computation cannot be performed in real time. A more 
efficient scheme is to have the codewords arranged in a tree \cite{CLG89}. 

The asymptotic behavior of quantization error, assuming optimal quantizers
and under various conditions on $P$, has been studied in great detail. A nice
overview is presented in the recent monograph of Graf and Luschgy \cite{GL00}.
The rates obtained for $k$-optimal quantizers are generally of the form $k^{-2/D}$.
There is also work on the special case of data that lie {\it exactly} on a 
manifold, and whose distribution is within some constant factor of uniform;
in such cases, rates of the form $k^{-2/d}$ are obtained, where $d$ is the
dimension of the manifold. Our setting is considerably more general than
this: we do not assume optimal quantization (which is NP-hard), we have 
a broad notion of intrinsic dimension that allows points to merely be close
to a manifold rather than on it, and we make no other assumptions about the
distribution $P$.

\subsubsection*{Compressed sensing}

The field of compressed sensing has grown out of the surprising realization
that high-dimensional sparse data can be accurately reconstructed from just
a few random projections \cite{CT06,D06}. The central premise of this research 
area is that the original data thus never even needs to be collected: all one 
ever sees are the random projections.

RP trees are similar in spirit and entirely compatible with this viewpoint.
Theorem~\ref{thm:main2} holds even if the random projections are forced to be 
the same across each entire level of the tree. For a tree of depth $k$, this 
means only $k$ random projections are ever needed, and these can be computed 
beforehand (the split-by-distance can be reworked to operate in the projected
space rather than the high-dimensional space). The data are not accessed in 
any other way.

\section{An RP tree adapts to intrinsic dimension}

An RP tree has two varieties of split. If a cell $C$ has much larger diameter
than average-diameter (average interpoint distance), then it is split 
according to the distances of points from the mean. Otherwise, a random projection 
is used.

The first type of split is particularly easy to analyze.

\subsection{Splitting by distance from the mean}

This option is invoked when the points in the current cell, call them $S$,
satisfy $\Delta^2(S) > c \Delta_A^2(S)$; recall that $\Delta(S)$ is the 
diameter of $S$ while $\Delta_A^2(S)$ is the average interpoint distance.

\begin{lemma}
Suppose that $\Delta^2(S) > c \Delta_A^2(S)$. Let $S_1$ denote the points in $S$ 
whose distance to $\mean(S)$ is less than or equal to the median distance, 
and let $S_2$ be the remaining points. 
Then the expected squared diameter after the split is
$$ \frac{|S_1|}{|S|} \Delta^2(S_1) + \frac{|S_2|}{|S|} \Delta^2(S_2) 
\ \leq \ 
\left( \frac{1}{2} + \frac{2}{c} \right) \Delta^2(S) .$$
\label{lemma:spherical-split}
\end{lemma}

The proof of this lemma is deferred to the Appendix, as are most of the other 
proofs in this paper.

\subsection{Splitting by projection: proof outline}

Suppose the current cell contains a set of points $S \subset \R^D$ for
which $\Delta^2(S) \leq c \Delta_A^2(S)$. We will show that a split by 
projection has a constant probability of reducing the average squared 
diameter $\Delta_A^2(S)$ by $\Omega(\Delta_A^2(S)/d)$. Our proof has 
three parts:
\begin{enumerate}
\item[I.] Suppose $S$ is split into $S_1$ and $S_2$, with means $\mu_1$ and
$\mu_2$. Then the reduction in average diameter can be expressed in a 
remarkably simple form, as a multiple of $\|\mu_1 - \mu_2\|^2$. 
\item[II.] Next, we give a lower bound on the distance between the 
{\it projected} means, $(\wmu_1 - \wmu_2)^2$. We show that the distribution
of the projected points is subgaussian with variance $O(\Delta_A^2(S)/D)$. 
This well-behavedness implies that 
$(\wmu_1 - \wmu_2)^2 = \Omega(\Delta_A^2(S)/D)$.
\item[III.] We finish by showing that, approximately, 
$\|\mu_1 - \mu_2\|^2 \geq (D/d) (\wmu_1 - \wmu_2)^2$.
This is because $\mu_1 - \mu_2$ lies close to the subspace spanned by the 
top $d$ eigenvectors of the covariance matrix of $S$; and with high probability, 
{\it every} vector in this subspace shrinks by  $O(\sqrt{d/D})$ 
when projected on a random line.
\end{enumerate}
We now tackle these three parts of the proof in order.

\subsection{Quantifying the reduction in average diameter}

The average squared diameter $\Delta_A^2(S)$ has certain reformulations
that make it convenient to work with.
These properties are consequences of the following two observations,
the first of which the reader may recognize as a standard 
``bias-variance'' decomposition of statistics.
\begin{lemma}
Let $X,Y$ be independent and identically distributed random variables
in $\R^n$, and let $z \in \R^n$ be any fixed vector.
\begin{enumerate}
\item[(a)] $ \E \left[ \|X - z\|^2 \right] = \E \left[ \| X - \E X\|^2 \right] + \| z - \E X\|^2$.
\item[(b)] $ \E \left[ \|X - Y\|^2 \right] = 2 \, \E \left[ \| X - \E X\|^2 \right] $.
\end{enumerate}
\label{lem:bias-var}
\end{lemma}
\begin{proof}
Part (a) is immediate when both sides are expanded. For (b), we use
part (a) to assert that for any fixed $y$, we have
$ \E \left[ \|X - y\|^2 \right] = \E \left[ \|X - \E X\|^2 \right] + \|y - \E X\|^2$.
We then take expectation over $Y = y$.
\end{proof}
This can be used to show that the averaged squared diameter, $\Delta_A^2(S)$, 
is twice the average squared distance of points in $S$ from their mean. 
\begin{cor}
The average squared diameter of a set $S$ can also be written as:
$$ \Delta_A^2(S) = \frac{2}{|S|} \sum_{x \in S} \|x - \mean(S)\|^2 .$$
\label{cor:defn-da}
\end{cor}
\begin{proof}
$\Delta_A^2(S)$ is simply $\E \left[ \|X - Y\|^2 \right]$, when $X,Y$ are 
i.i.d.\ draws from the uniform distribution over $S$.
\end{proof}

At each successive level of the tree, the current cell is split into two,
either by a random projection or according to distance from the mean. 
Suppose the points in the current cell are $S$, and that they are split
into sets $S_1$ and $S_2$. It is obvious that the expected diameter is
nonincreasing:
$$ \Delta(S) \ \geq \ \frac{|S_1|}{|S|} \Delta(S_1) + \frac{|S_2|}{|S|} \Delta(S_2) .$$
This is also true of the expected average diameter. In fact, we can
precisely characterize how much it decreases on account of the split.
\begin{lemma}
Suppose set $S$ is partitioned (in any manner) into $S_1$ and $S_2$.
Then 
\begin{eqnarray*}
\Delta_A^2(S) - \left\{ \frac{|S_1|}{|S|} \Delta_A^2(S_1) + \frac{|S_2|}{|S|} \Delta_A^2(S_2) \right\}
& = & \frac{2 |S_1| \cdot |S_2|}{|S|^2} \, \| \mean(S_1) - \mean(S_2) \|^2 .
\end{eqnarray*}
\label{lemma:decrease-in-DA}
\end{lemma}
This completes part I of the proof outline.

\subsection{Properties of random projections}

Our quantization scheme depends heavily upon certain regularity properties 
of random projections. We now review these properties, which are critical for
parts II and III of our proof.

The most obvious way to pick a random projection from $\R^D$ to $\R$ is to 
choose a projection direction $u$ uniformly at random from the surface of the 
unit sphere $S^{D-1}$, and to send $x \mapsto u \cdot x$.

Another common option is to select the projection vector from a multivariate
Gaussian distribution, $u \sim N(0, (1/D) I_D)$. This gives almost the same
distribution as before, and is slightly easier to work with in terms of the
algorithm and analysis. We will therefore use this type of projection, bearing
in mind that all proofs carry over to the other variety as well, with slight 
changes in constants.

The key property of a random projection from $\R^D$ to $\R$ is that it 
approximately preserves the lengths of vectors, modulo a scaling factor of 
$\sqrt{D}$. This is summarized in the lemma below.

\begin{lemma}
Fix any $x \in \R^D$. Pick a random vector $U \sim N(0, (1/D) I_D)$. Then
for any $\alpha, \beta > 0$:
\begin{enumerate}
\item[(a)] $\P \left[ |U \cdot x| \leq \alpha \cdot \frac{\|x\|}{\sqrt{D}} \right] 
\ \leq \ \sqrt{\frac{2}{\pi}}\, \alpha$
\item[(b)] $\P \left[ |U \cdot x| \geq  \beta \cdot \frac{\|x\|}{\sqrt{D}} \right]  
\ \leq \ \frac{2}{\beta} e^{-\beta^2/2}$
\end{enumerate}
\label{lemma:normal}
\end{lemma}


Lemma~\ref{lemma:normal} applies to any individual vector. Thus it also applies, 
in expectation, to a vector chosen at random from a set $S \subset \R^D$. Applying 
Markov's inequality, we can then conclude that when $S$ is projected onto a random 
direction, {\it most} of the projected points will be close together, in a 
{\it central interval} of size $O(\Delta(S)/\sqrt{D})$. 

\begin{lemma}
Suppose $S \subset \R^D$ lies within some ball $B(x_0, \Delta)$. Pick any 
$0 < \delta, \epsilon \leq 1$ such that $\delta \epsilon \leq 1/e^2$.
Let $\nu$ be any measure on $S$. Then with probability $> 1 - \delta$ over 
the choice of random projection $U$ onto $\R$, all but an $\epsilon$ fraction of 
$U \cdot S$ (measured according to $\nu$) lies within distance
$ \sqrt{2 \ln \frac{1}{\delta \epsilon}} \cdot \frac{\Delta}{\sqrt{D}}$
of $U \cdot x_0$.
\label{lemma:near-x0}
\end{lemma}

As a corollary, the median of the projected points must also lie within this
central interval.

\begin{cor}
Under the hypotheses of Lemma~\ref{lemma:near-x0}, for any $0 < \delta < 2/e^2$, the
following holds with probability at least $1-\delta$ over the choice of projection:
$$ | \med(U \cdot S) - U \cdot x_0 | 
\ \ \leq \ \ 
\frac{\Delta}{\sqrt{D}} \cdot \sqrt{2 \ln \frac{2}{\delta}}.
$$
\label{cor:median}
\end{cor}
\begin{proof}
Let $\nu$ be the uniform distribution over $S$ and use $\epsilon = 1/2$.
\end{proof}

Finally, we examine what happens when the set $S$ is a $d$-dimensional subspace
of $\R^D$. Lemma~\ref{lemma:normal} tells us that the projection of any {\it specific} 
vector $x \in S$ is unlikely to have length too much greater than $\|x\|/\sqrt{D}$,
with high probability. A slightly weaker bound can be shown to hold for all of $S$ 
simultaneously; the proof technique has appeared before in several contexts, including 
\cite{M71,BDDW08}.
\begin{lemma}
There exists a constant $\kappa_1$ with the following property.
Fix any $\delta > 0$ and any $d$-dimensional subspace $H \subset \R^D$. 
Pick a random projection $U \sim N(0,(1/D)I_D)$. Then with probability
at least $1- \delta$ over the choice of $U$,
$$ \sup_{x \in H} \frac{|x \cdot U|^2}{\|x\|^2} \ \leq \ \kappa_1 \cdot 
\frac{d + \ln 1/\delta}{D} .$$
\label{lemma:projected-subspace}
\end{lemma}
\begin{proof}
It is enough to show that the inequality holds for
$S = H \cap \mbox{(surface of the unit sphere in $\R^D$)}$. Let $N$ be any 
$(1/2)$-cover of this set; it is possible to achieve 
$|N| \leq 10^d$ \cite{M02}. Apply Lemma~\ref{lemma:normal}, along with 
a union bound, to conclude that with probability at least $1-\delta$ over the 
choice of projection $U$,
$$ \sup_{x \in N} |x \cdot U|^2 
\ \leq \ 2 \cdot \frac{\ln |N| + \ln 1/\delta}{D} .$$ 
Now, define $C$ by
$$ C \ = \ \sup_{x \in S} \left( |x \cdot U|^2 \cdot \frac{D}{\ln |N| + \ln 1/\delta} \right) .$$ 
We'll complete the proof by showing $C \leq 8$. To this end, pick the $x^* \in S$ for
which the supremum is realized (note $S$ is compact), and choose $y \in N$ whose 
distance to $x^*$ is at most $1/2$. Then, 
\begin{eqnarray*}
|x^* \cdot U| 
& \leq & |y \cdot U| + | (x^* - y) \cdot U| \\
& \leq & \sqrt{\frac{\ln |N| + \ln 1/\delta}{D}} \left( \sqrt{2} + \frac{1}{2}\sqrt{C} \right)
\end{eqnarray*}
From the definition of $x^*$, it follows that $\sqrt{C} \leq \sqrt{2} + \sqrt{C}/2$ and thus $C \leq 8$.
\end{proof}

\subsection{Properties of the projected data}

Projection from $\R^D$ into $\R^1$ shrinks the average squared
diameter of a data set by roughly $D$.
To see this, we start with the fact that when a data 
set with covariance  $A$ is projected onto a vector $U$, the 
projected data have variance $U^T A U$. We now show that for 
random $U$, such quadratic forms are concentrated about their 
expected values.
\begin{lemma}
Suppose $A$ is an $n \times n$ positive semidefinite matrix, and
$U \sim N(0, (1/n) I_n)$. Then for any $\alpha, \beta > 0$:
\begin{enumerate}
\item[(a)] $\P[ U^T A U < \alpha \cdot \E[U^T A U]] \leq e^{-((1/2) - \alpha)/2}$, and
\item[(b)] $\P[ U^T A U > \beta \cdot \E[U^T A U]] \leq e^{-(\beta-2)/4}$.
\end{enumerate}
\label{lemma:projected-variance}
\end{lemma}

\begin{lemma}
Pick $U \sim N(0,(1/D)I_D)$. Then for any $S \subset \R^D$, with probability at 
least $1/10$, the projection of $S$ onto $U$ has average squared diameter
$$ \Delta_A^2(S \cdot U) \geq \frac{\Delta_A^2(S)}{4D}.$$
\label{lemma:variance2}
\end{lemma}
\begin{proof}
By Corollary~\ref{cor:defn-da},
$$
\Delta_A^2(S \cdot U)
\ = \ 
\frac{2}{|S|} \sum_{x \in S} ((x - \mean(S)) \cdot U)^2 
\ = \ 
2 U^T \cov(S) U.
$$
where $\cov(S)$ is the covariance of data set $S$. This quadratic term has
expectation (over choice of $U$)
\begin{eqnarray*}
\E[2 U^T \cov(S) U] 
& = &
2 \sum_{i,j} \E[U_i U_j] \cov(S)_{ij} \\
& = &  
\frac{2}{D} \sum_i \cov(S)_{ii}
\ = \ 
\frac{\Delta_A^2(S)}{D} .
\end{eqnarray*}
Lemma~\ref{lemma:projected-variance}(a) then bounds the probability that 
it is much smaller than its expected value.
\end{proof}

Next, we examine the overall distribution of the projected points.
When $S \subset \R^D$ has diameter $\Delta$, its projection into the 
line can have diameter upto $\Delta$, but as we saw in 
Lemma~\ref{lemma:near-x0}, most of it will lie within a central 
interval of size $O(\Delta/\sqrt{D})$. What can be said about points 
that fall outside this interval?
\begin{lemma}
Suppose $S \subset B(0, \Delta) \subset \R^D$. Pick any $\delta > 0$ and
choose $U \sim N(0,(1/D)I_D)$. Then with probability at least $1- \delta$ 
over the choice of $U$, the projection $S \cdot U = \{x \cdot U: x \in S\}$ 
satisfies the following property for all positive integers $k$.
\begin{quote}
The fraction of points outside the interval
$\left(- \frac{k\Delta}{\sqrt{D}}, +\frac{k\Delta}{\sqrt{D}} \right)$ 
is at most $\frac{2^k}{\delta} \cdot e^{-k^2/2}$.
\end{quote}
\label{lemma:distribution}
\end{lemma}
\begin{proof}
This follows by applying Lemma~\ref{lemma:near-x0} for each positive
integer $k$ (with corresponding failure probability $\delta/2^k$), and 
then taking a union bound.
\end{proof}

\subsection{Distance between the projected means}

We are dealing with the case when $\Delta^2(S) \leq c \cdot \Delta_A^2(S)$, 
that is, the diameter of set $S$ is at most a constant factor times 
the average interpoint distance. If $S$ is projected onto a random direction,
the projected points will have variance about $\Delta_A^2(S)/D$, by 
Lemma~\ref{lemma:variance2}; and by Lemma~\ref{lemma:distribution}, it 
isn't too far from the truth to think of these points as having roughly 
a Gaussian distribution. Thus, if the projected points are split into two 
groups at the mean, we would expect the means of these two groups to be 
separated by a distance of about $\Delta_A(S)/\sqrt{D}$. Indeed, this is 
the case. The same holds if we split at the median, which isn't all that
different from the mean for close-to-Gaussian distributions.

\begin{lemma}
There is a constant $\kappa_2$ for which the following holds.
Pick any $0 < \delta < 1/16c$. Pick $U \sim N(0, (1/D) I_D)$ and split $S$ into two 
pieces:
$$ S_1 = \{x \in S: x \cdot U < s\} 
\mbox{\ \ and \ \ } 
S_2 = \{x \in S: x \cdot U \geq s\} ,$$
where $s$ is either $\mean(S \cdot U)$ or $\mbox{median}(S \cdot U)$.
Write $p = |S_1|/|S|$, and let $\wmu_1$ and $\wmu_2$ denote the means of 
$S_1 \cdot U$ and $S_2 \cdot U$, respectively. Then with probability at least 
$1/10 - \delta$,
$$ 
(\wmu_2 - \wmu_1)^2 
\ \geq \  
\kappa_2 \cdot \frac{1}{(p(1-p))^2} \cdot
\frac{\Delta_A^2(S)}{D} \cdot
\frac{1}{c \, \log (1/\delta)}  .$$
\label{lemma:dist-btwn-means}
\end{lemma}

\begin{proof}
Let the random variable $\wX$ denote a uniform-random draw from the projected
points $S \cdot U$. Without loss of generality $\mean(S) = 0$, so that
$\E \wX = 0$ and thus $p \wmu_1 + (1-p) \wmu_2 = 0$. Rearranging, we get 
$\wmu_1 = -(1-p) (\wmu_2 - \wmu_1)$ and $\wmu_2 = p (\wmu_2 - \wmu_1)$.

We already know from Lemma~\ref{lemma:variance2} (and Corollary~\ref{cor:defn-da}) 
that with probability at least $1/10$, the variance of the projected points is 
significant: $\var(\wX) \geq \Delta_A^2(S)/8D$. We'll show this implies a similar
lower bound on $(\wmu_2 - \wmu_1)^2$. 

Using ${\bf 1}(\cdot)$ to denote $0-1$ indicator variables,
\begin{eqnarray*}
\var(\wX)
& \leq &
\E [ (\wX-s)^2 ] \\ 
& \leq &
\E [2t |\wX - s| + (|\wX - s| - t)^2 \cdot {\bf 1}(|\wX - s| \geq t)]
\end{eqnarray*}
for any $t > 0$.
This is a convenient formulation since the linear term gives us $\wmu_2 - \wmu_1$:
\begin{eqnarray*}
\E [2t |\wX - s|]  
& = &
2t (p (s - \wmu_1) + (1-p) (\wmu_2 - s)) \\
& = &
4t \cdot p(1-p) \cdot (\wmu_2 - \wmu_1) + 2ts(2p-1).
\end{eqnarray*}
The last term vanishes since the split is either at the mean of the projected points, in
which case $s=0$, or at the median, in which case $p = 1/2$.

Next, we'll choose
$$ t \ \ = \ \  t_o \frac{\Delta(S)}{\sqrt{D}} \cdot \sqrt{\log \frac{1}{\delta}} $$
for some suitable constant $t_o$, so that the quadratic term in $\var(\wX)$ can be bounded using 
Lemma~\ref{lemma:distribution} and Corollary~\ref{cor:median}: with probability at 
least $1-\delta$,
$$ E[(|\wX|-t)^2 \cdot {\bf 1}(|\wX| \geq t)] \ \ \leq \ \ \delta \cdot \frac{\Delta^2(S)}{D}$$ 
(this is a simple integration). Putting the pieces together, we have
$$ 
\frac{\Delta_A^2(S)}{8D} 
\ \ \leq \ \   
\var(\wX) 
\ \ \leq \ \   
4t \cdot p(1-p) \cdot (\wmu_2 - \wmu_1) + \delta \cdot \frac{\Delta^2(S)}{D} .
$$
The result now follows immediately by algebraic manipulation, using the
relation $\Delta^2(S) \leq c \Delta_A^2(S)$.
\end{proof}

\subsection{Distance between the high-dimensional means}

Split $S$ into two pieces as in the setting of Lemma~\ref{lemma:dist-btwn-means}, 
and let $\mu_1$ and $\mu_2$ denote the means of $S_1$ and $S_2$, respectively. We
already have a lower bound on the distance between the projected means, 
$\wmu_2 - \wmu_1$; we will now show that $\|\mu_2 - \mu_1\|$ is larger than this
by a factor of about $\sqrt{D/d}$. The main technical difficulty here is the 
dependence between the $\mu_i$ and the projection $U$. Incidentally, this is the 
only part of the entire argument that exploits intrinsic dimensionality.

\begin{lemma}
There exists a constant $\kappa_3$ with the following property.
Suppose set $S \subset \R^D$ is such that the top $d$ eigenvalues of 
$\cov(S)$ account for more than $1-\epsilon$ of its trace. Pick a random vector 
$U \sim N(0, (1/D) I_D)$, and split $S$ into two pieces, $S_1$ and $S_2$, in any 
fashion (which may depend upon $U$). Let $p = |S_1|/|S|$. Let $\mu_1$ and 
$\mu_2$ be the means of $S_1$ and $S_2$, and let $\wmu_1$ and $\wmu_2$ be the 
means of $S_1 \cdot U$ and $S_2 \cdot U$.

Then, for any $\delta > 0$, with probability at least $1-\delta$ over the choice of $U$,
$$ \|\mu_2 - \mu_1\|^2 \ \geq \ \frac{\kappa_3 D}{d + \ln 1/\delta} 
\left( (\wmu_2 - \wmu_1)^2 - \frac{4}{p(1-p)} \frac{\epsilon \Delta_A^2(S)}{\delta D} \right) .$$
\label{lemma:high-vs-low}
\end{lemma}
\begin{proof}
Assume without loss of generality that $S$ has zero mean. 
Let $H$ denote the subspace spanned by the top $d$ eigenvectors of the covariance matrix 
of $S$, and let $H^\bot$ be its orthogonal subspace. Write any point $x \in \R^D$ as 
$x_H + x_\bot$, where each component is seen as a vector in $\R^D$ that lies in the 
respective subspace.

Pick the random vector $U$; with probability $\geq 1 - \delta$ it satisfies the following
two properties.

\noindent
Property 1: For some constant $\kappa' > 0$, for every $x \in \R^D$ 
$$ |x_H \cdot U|^2 
\ \leq \ 
\|x_H\|^2 \cdot \kappa' \cdot \frac{d + \ln 1/\delta}{D}
\ \leq \ 
\|x\|^2 \cdot \kappa' \cdot \frac{d + \ln 1/\delta}{D} .$$
This holds (with probability $1- \delta/2$) by Lemma~\ref{lemma:projected-subspace}.

\noindent
Property 2: Letting $X$ denote a uniform-random draw from $S$, we have
\begin{eqnarray*}
\E_X [ (X_\bot \cdot U)^2 ] 
& \leq &
\frac{2}{\delta} \cdot \E_U \E_X [(X_\bot \cdot U)^2] \\
& = &
\frac{2}{\delta} \cdot \E_X \E_U [(X_\bot \cdot U)^2] \\
& = & 
\frac{2}{\delta D} \cdot \E_X [\|X_\bot\|^2]
\ \leq \ 
\frac{\epsilon \Delta_A^2(S)}{\delta D}.
\end{eqnarray*}
The first step is Markov's inequality, and holds with probability $1-\delta/2$.
The last inequality comes from the local covariance condition.

So assume the two properties hold. Writing 
$\mu_2 - \mu_1$ as $(\mu_{2H} - \mu_{1H}) + (\mu_{2\bot} - \mu_{1\bot})$,
\begin{eqnarray*}
(\wmu_2 - \wmu_1)^2 
& = &
((\mu_{2H} - \mu_{1H}) \cdot U + (\mu_{2\bot} - \mu_{1\bot}) \cdot U)^2 \\
& \leq &
2 ((\mu_{2H} - \mu_{1H}) \cdot U)^2 + 2 ((\mu_{2\bot} - \mu_{1\bot}) \cdot U)^2 .
\end{eqnarray*}
The first term can be bounded by Property 1: 
$$
((\mu_{2H} - \mu_{1H}) \cdot U)^2 
\ \leq \ 
\| \mu_2 - \mu_1 \|^2 \cdot \kappa' \cdot \frac{d + \ln 1/\delta}{D}.
$$
For the second term, let $\E_X$ denote expectation over $X$ chosen uniformly at 
random from $S$. Then
\begin{eqnarray*}
((\mu_{2\bot} - \mu_{1\bot}) \cdot U)^2 
& \leq &
2 (\mu_{2\bot} \cdot U)^2 + 2 (\mu_{1\bot} \cdot U)^2 \\
& = &
2 (\E_X [X_\bot \cdot U\ |\ X \in S_2])^2 + 2 (\E_X [X_\bot \cdot U\ |\ X \in S_1])^2 \\
& \leq &
2 \E_X [(X_\bot \cdot U)^2 \ |\ X \in S_2] + 2 \E_X [(X_\bot \cdot U)^2 \ |\ X \in S_1] \\
& \leq & 
\frac{2}{1-p} \cdot \E_X [(X_\bot \cdot U)^2] + \frac{2}{p} \cdot \E_X [(X_\bot \cdot U)^2] \\
&  =  & 
\frac{2}{p(1-p)} \E_X [ (X_\bot \cdot U)^2 ]
\ \leq \ 
\frac{2}{p(1-p)} \cdot \frac{\epsilon \Delta_A^2(S)}{\delta D}.
\end{eqnarray*}
by Property 2. The lemma follows by putting the various pieces together.
\end{proof}
We can now finish off the proof of Theorem~\ref{thm:main2}.
\begin{thm}
Fix any $\epsilon \leq O(1/c)$. 
Suppose set $S \subset \R^D$ has the property that the top $d$ eigenvalues of $\cov(S)$
account for more than $1-\epsilon$ of its trace. Pick a random vector $U \sim N(0,(1/D)I_D)$
and split $S$ into two parts,
$$ S_1 = \{x \in S: x \cdot U < s\} 
\mbox{\ \ and \ \ } 
S_2 = \{x \in S: x \cdot U \geq s\} ,$$
where $s$ is either $\mean(S \cdot U)$ or $\mbox{median}(S \cdot U)$.
Then with probability $\Omega(1)$, the expected average diameter shrinks by 
$\Omega(\Delta_A^2(S)/cd)$.
\end{thm}
\begin{proof}
By Lemma~\ref{lemma:decrease-in-DA}, the reduction in expected average diameter is
\begin{eqnarray*}
\Delta_A^2(S) - \left\{ \frac{|S_1|}{|S|} \Delta_A^2(S_1) + \frac{|S_2|}{|S|} \Delta_A^2(S_2) \right\} 
& = & 
\frac{2 |S_1| \cdot |S_2|}{|S|^2} \, \| \mean(S_1) - \mean(S_2) \|^2 ,
\end{eqnarray*}
or $2p(1-p) \|\mu_1 - \mu_2\|^2$ in the language of Lemmas~\ref{lemma:dist-btwn-means} and \ref{lemma:high-vs-low}. The rest follows from those two lemmas.
\end{proof}

\subsubsection*{Acknowledgements}

Dasgupta acknowledges the support of the National Science Foundation under 
grants IIS-0347646 and IIS-0713540.

\bibliographystyle{plain}
\bibliography{manifold}

\begin{thebibliography}{10}

\bibitem{BDDW08}
R.~Baraniuk, M.~Davenport, R.~DeVore, and M.~Wakin.
\newblock A simple proof of the restricted isometry property for random
  matrices.
\newblock {\em Constructive Approximation}, 2008.

\bibitem{BN03}
M.~Belkin and P.~Niyogi.
\newblock Laplacian eigenmaps for dimensionality reduction and data
  representation.
\newblock {\em Neural Computation}, 15(6):1373--1396, 2003.

\bibitem{CT06}
E.~Candes and T.~Tao.
\newblock Near optimal signal recovery from random projections: universal
  encoding strategies?
\newblock {\em IEEE Transactions on Information Theory}, 52(12):5406--5425,
  2006.

\bibitem{CLG89}
P.A. Chou, T.~Lookabaugh, and R.M. Gray.
\newblock Optimal pruning with applications to tree-structured source coding
  and modeling.
\newblock {\em IEEE Transactions on Information Theory}, 35(2):299--315, 1989.

\bibitem{D06}
D.~Donoho.
\newblock Compressed sensing.
\newblock {\em IEEE Transactions on Information Theory}, 52(4):1289--1306,
  2006.

\bibitem{DFKVV04}
P.~Drineas, A.~Frieze, R.~Kannan, S.~Vempala, and V.~Vinay.
\newblock Clustering large graphs via the singular value decomposition.
\newblock {\em Machine Learning}, 56:9--33, 2004.

\bibitem{D95}
R.~Durrett.
\newblock {\em Probability: Theory and Examples}.
\newblock Duxbury, second edition, 1995.

\bibitem{GL00}
S.~Graf and H.~Luschgy.
\newblock {\em Foundations of quantization for probability distributions}.
\newblock Springer, 2000.

\bibitem{GN98}
R.M. Gray and D.L. Neuhoff.
\newblock Quantization.
\newblock {\em IEEE Transactions on Information Theory}, 44(6):2325--2383,
  1998.

\bibitem{KW78}
J.B. Kruskal and M.~Wish.
\newblock {\em Multidimensional Scaling}.
\newblock Sage University Paper series on Quantitative Application in the
  Social Sciences, 07-011. 1978.

\bibitem{L82}
S.P. Lloyd.
\newblock Least squares quantization in pcm.
\newblock {\em IEEE Transactions on Information Theory}, 28(2):129--137, 1982.

\bibitem{MacQ67}
J.B. MacQueen.
\newblock Some methods for classification and analysis of multivariate
  observations.
\newblock In {\em Proceedings of Fifth Berkeley Symposium on Mathematical
  Statistics and Probability}, volume~1, pages 281--297. University of
  California Press, 1967.

\bibitem{M02}
J.~Matousek.
\newblock {\em Lectures on Discrete Geometry}.
\newblock Springer, 2002.

\bibitem{M71}
V.D. Milman.
\newblock A new proof of the theorem of a. dvoretsky on sections of convex
  bodies.
\newblock {\em Functional Analysis and its Applications}, 5(4):28--37, 1971.

\bibitem{P82}
D.~Pollard.
\newblock Quantization and the method of $k$-means.
\newblock {\em IEEE Transactions on Information Theory}, 28:199--205, 1982.

\bibitem{RS00}
S.T. Roweis and L.K. Saul.
\newblock Nonlinear dimensionality reduction by locally linear embedding.
\newblock {\em Science}, (290):2323--2326, 2000.

\bibitem{S38}
I.J. Schoenberg.
\newblock Metric spaces and positive definite functions.
\newblock {\em Transactions of the American Mathematical Society}, 44:522--553,
  1938.

\bibitem{TdSL00}
J.~Tenenbaum, V.~de~Silva, and J.~Langford.
\newblock A global geometric framework for nonlinear dimensionality reduction.
\newblock {\em Science}, 290(5500):2319--2323, 2000.

\end{thebibliography}

\section{Appendix I: Proofs of main theorem}

\subsection{Proof of Lemma~\ref{lemma:normal}}

Since $U$ has a Gaussian distribution, and any linear combination of independent 
Gaussians is a Gaussian, it follows that the projection $U \cdot x$ is also Gaussian. 
Its mean and variance are easily seen to be zero and $\|x\|^2/D$, respectively. 
Therefore, writing
$$ Z \ = \ \frac{\sqrt{D}}{\|x\|}\, (U \cdot x) $$
we have that $Z \sim N(0,1)$. The bounds stated in the lemma now follow from properties
of the standard normal. In particular, $N(0,1)$ is roughly flat in the range $[-1,1]$
and then drops off rapidly; the two cases in the lemma statement correspond to these 
two regimes.

The highest density achieved by the standard normal is $1/\sqrt{2\pi}$. Thus the 
probability mass it assigns to the interval $[-\alpha,\alpha]$ is at most 
$2\alpha/\sqrt{2 \pi}$; this takes care of (a). For (b), we use 
a standard tail bound for the normal, $\P(|Z| \geq \beta) \leq (2/\beta)e^{-\beta^2/2}$; 
see, for instance, page 7 of \cite{D95}.

\subsection{Proof of Lemma~\ref{lemma:near-x0}}

Set $c = \sqrt{2 \ln 1/(\delta\epsilon)} \geq 2$.

Fix any point $x$, and randomly choose a projection $U$. Let $\wx = U \cdot x$ (and 
likewise, let $\wS = U \cdot S$). What is the chance that
$\wx$ lands far from $\wx_0$? Define the bad event to be 
$F_x = {\bf 1}(|\wx - \wx_0| \geq c \Delta/\sqrt{D})$. 
By Lemma~\ref{lemma:normal}(b), we have
$$ \E_U [F_x] 
\ \leq \ 
\P_U \left[ |\wx - \wx_0| \geq c \cdot \frac{\|x - x_0\|}{\sqrt{D}} \right] 
\ \leq \ 
\frac{2}{c} \, e^{-c^2/2}
\ \leq \ 
\delta \epsilon .$$
Since this holds for any $x \in S$, it also holds in expectation over $x$ drawn
from $\nu$. We are interested in bounding the probability (over the choice of
$U$) that more than an $\epsilon$ fraction of $\nu$ falls far from $\wx_0$. 
Using Markov's inequality and then Fubini's theorem, we have
$$ \P_U \left[ \E_\mu [F_x] \geq \epsilon \right] 
\ \leq \  
\frac{\E_U [\E_\mu [F_x]]}{\epsilon}
\ = \ 
\frac{\E_\mu [\E_U [F_x]]}{\epsilon}
\ \leq \  
\delta,
$$
as claimed.

\subsection{Proof of Lemma~\ref{lemma:spherical-split}}

Let random variable $X$ be distributed uniformly over $S$. Then
$$ \P \left[ \|X - \E X \|^2 \ \geq \ \med(\|X - \E X \|^2) \right] \ \geq \ \frac{1}{2} $$
by definition of median, so $\E \left[ \|X - \E X \|^2 \right] \geq \med(\|X - \E X \|^2)/2$.
It follows from Corollary~\ref{cor:defn-da} that
$$ \med(\|X - \E X \|^2) \leq 2 \E \left[ \|X - \E X \|^2 \right] = \Delta_A^2(S).$$

Set $S_1$ has squared diameter 
$\Delta^2(S_1) \leq (2 \, \med(\|X - \E X \|))^2 \leq 4 \Delta_A^2(S)$.
Meanwhile, $S_2$ has squared diameter at most $\Delta^2(S)$. Therefore,
$$ 
\frac{|S_1|}{|S|} \Delta^2(S_1) + \frac{|S_2|}{|S|} \Delta^2(S_2) 
\ \leq \ 
\frac{1}{2} \cdot 4 \Delta_A^2(S) + \frac{1}{2} \Delta^2(S)$$
and the lemma follows by using $\Delta^2(S) > c \Delta_A^2(S)$.

\subsection{Proof of Lemma~\ref{lemma:projected-variance}}

This follows by examining the moment-generating function of $U^T A U$.
Since the distribution of $U$ is spherically symmetric, we can work in 
the eigenbasis of $A$ and assume without loss of generality that 
$A = \mbox{diag}(a_1, \ldots, a_n)$, where $a_1, \ldots, a_n$ are the 
eigenvalues. Moreover, for convenience we take $\sum a_i = 1$.

Let $U_1, \ldots, U_n$ denote the individual coordinates of $U$. We can
rewrite them as $U_i = Z_i/\sqrt{n}$, where $Z_1,\ldots,Z_n$ are 
i.i.d.\ standard normal random variables. Thus
$$ U^T A U \ = \ \sum_i a_i U_i^2 \ = \ \frac{1}{n} \sum_i a_i Z_i^2 .$$
This tells us immediately that $\E [U^T A U] = 1/n$.

We use Chernoff's bounding method for both parts. For (a), for any $t > 0$,
\begin{eqnarray*}
\P \left[ U^T A U < \alpha \cdot \E[U^T A U] \right] 
& = & 
\P \left[ \sum_i a_i Z_i^2 < \alpha \right] 
\ \  =  \ \  
\P \left[ e^{-t \sum_i a_i Z_i^2} > e^{-t \alpha} \right] \\
& \leq &
\frac{\E\left[ e^{- t\sum_i a_i Z_i^2} \right]}{e^{-t\alpha}} 
\ \  = \ \  
e^{t \alpha} \prod_i \E \left[ e^{-t a_i Z_i^2} \right] \\
& = & 
e^{t \alpha} \prod_i \left( \frac{1}{1 + 2ta_i} \right)^{1/2} 
\end{eqnarray*}
and the rest follows by using $t = 1/2$ along with the inequality
$1/(1+x) \leq e^{-x/2}$ for $0 < x \leq 1$. Similarly for (b),
for $0 < t < 1/2$,
\begin{eqnarray*}
\P \left[ U^T A U > \beta \cdot \E[U^T A U] \right]
& = & 
\P \left[ \sum_i a_i Z_i^2 > \beta \right] 
\ \  =  \ \  
\P \left[ e^{t \sum_i a_i Z_i^2} > e^{t \beta} \right] \\
& \leq &
\frac{\E\left[e^{t\sum_i a_i Z_i^2} \right]}{e^{t\beta}}
\ \  =  \ \  
e^{- t \beta} \prod_i \E \left[ e^{t a_i Z_i^2} \right] \\
& = & 
e^{- t \beta} \prod_i \left( \frac{1}{1 - 2ta_i} \right)^{1/2} 
\end{eqnarray*}
and it is adequate to choose $t= 1/4$ and invoke the inequality
$1/(1-x) \leq e^{2x}$ for $0 < x \leq 1/2$.

\subsection{Proof of Lemma~\ref{lemma:decrease-in-DA}}

Let $\mu, \mu_1, \mu_2$ denote the means of $S$, $S_1$, and $S_2$. 
Using Corollary~\ref{cor:defn-da} and Lemma~\ref{lem:bias-var}(a), 
we have
\begin{eqnarray*}
\lefteqn{\Delta_A^2(S) - \frac{|S_1|}{|S|} \Delta_A^2(S_1) - \frac{|S_2|}{|S|} \Delta_A^2(S_2)}\\
& = & 
\frac{2}{|S|} \sum_{S} \| x - \mu\|^2 
- \frac{|S_1|}{|S|} \cdot \frac{2}{|S_1|} \sum_{S_1} \|x - \mu_1 \|^2 
- \frac{|S_2|}{|S|} \cdot \frac{2}{|S_2|} \sum_{S_2} \|x - \mu_2 \|^2 \\
& = &
\frac{2}{|S|} 
\left\{ 
\sum_{S_1} \left( \|x - \mu\|^2 - \|x - \mu_1\|^2 \right) +
\sum_{S_2} \left( \|x - \mu\|^2 - \|x - \mu_2\|^2 \right)
\right\} \\
& = &
\frac{2 |S_1|}{|S|} \| \mu_1 - \mu \|^2 + \frac{2 |S_2|}{|S|} \| \mu_2 - \mu \|^2 .
\end{eqnarray*}
Writing $\mu$ as a weighted average of $\mu_1$ and $\mu_2$ then 
completes the proof.

\section{Appendix II: Hardness of $k$-means clustering}

\begin{quote}
{\sc $k$-means clustering}

{\it Input:} Set of points $x_1, \ldots, x_n \in \R^d$; integer $k$.

{\it Output:} A partition of the points into clusters $C_1, \ldots, C_k$,
along with a center $\mu_j$ for each cluster, so as to minimize
$$ \sum_{j=1}^k \sum_{i \in C_j} \|x_i - \mu_j \|^2 .$$
\end{quote}
(Here $\|\cdot\|$ is Euclidean distance.)
It can be checked that in any optimal solution, $\mu_j$ is the mean
of the points in $C_j$. Thus the $\{\mu_j\}$ can be removed entirely from
the formulation of the problem. From Lemma~\ref{lem:bias-var}(b),
$$ \sum_{i \in C_j} \|x_i - \mu_j \|^2 = \frac{1}{2|C_j|} \sum_{i, i' \in C_j} \|x_i - x_{i'} \|^2.$$
Therefore, the $k$-means cost function can equivalently be rewritten as
$$ \sum_{j=1}^k \frac{1}{2 |C_j|} \sum_{i, i' \in C_j} \|x_i - x_{i'} \|^2 .$$

We consider the specific case when $k$ is fixed to 2.
\begin{thm}
$2$-means clustering is an NP-hard optimization problem.
\label{thm:main}
\end{thm}
This was recently asserted in \cite{DFKVV04}, but the proof was flawed.
We establish hardness by a sequence of reductions. Our starting point is
a standard restriction of {\sc 3Sat} that is well known to be NP-complete.
\begin{quote}
{\sc 3Sat}

{\it Input:} A Boolean formula in 3CNF, where each clause has exactly three
literals and each variable appears at least twice.

{\it Output:} {\tt true} if formula is satisfiable, {\tt false} if not.
\end{quote}
By a standard reduction from {\sc 3Sat}, we show that a special case of 
{\sc not-all-equal 3Sat} is also hard. For completeness, the details are 
laid out in the next section.
\begin{quote}
{\sc NaeSat*}

{\it Input:} A Boolean formula $\phi(x_1, \ldots, x_n)$ in 3CNF, such that 
(i) every clause contains exactly three literals, and (ii) each pair of 
variables $x_i, x_j$ appears together in at most two clauses, once as 
either $\{x_i, x_j\}$ or $\{\overline{x}_i, \overline{x}_j\}$, and
once as either $\{\overline{x}_i, x_j\}$ or $\{x_i, \overline{x}_j\}$.

{\it Output:} {\tt true} if there exists an assignment in which 
each clause contains exactly one or two satisfied literals; {\tt false}
otherwise.
\end{quote}
Finally, we get to a generalization of {\sc 2-means}.
\begin{quote}
{\sc Generalized 2-means}

{\it Input:} An $n \times n$ matrix of interpoint distances $D_{ij}$.

{\it Output:} A partition of the points into two clusters $C_1$ and $C_2$,
so as to minimize
$$ \sum_{j=1}^2 \frac{1}{2 |C_j|} \sum_{i, i' \in C_j} D_{ii'} .$$
\end{quote}
We reduce {\sc NaeSat*} to {\sc Generalized 2-means}. For 
any input $\phi$ to {\sc NaeSat*}, we show how to efficiently produce  
a distance matrix $D(\phi)$ and a threshold $c(\phi)$ such that
$\phi$ satisfies {\sc NaeSat*} if and only if $D(\phi)$ admits a 
generalized 2-means clustering of cost $\leq c(\phi)$.

Thus {\sc Generalized 2-means clustering} is hard. To get back 
to {\sc 2-means} (and thus establish Theorem~\ref{thm:main}), we prove 
that the distance matrix $D(\phi)$ can 
in fact be realized by squared Euclidean distances. This existential 
fact is also constructive, because in such cases, the embedding can be 
obtained in cubic time by classical multidimensional scaling \cite{KW78}.

\subsection{Hardness of {\sc NaeSat*}}

Given an input $\phi(x_1, \ldots, x_n)$ to {\sc 3Sat}, we first 
construct an intermediate formula $\phi'$ that is satisfiable if and only 
if $\phi$ is, and additionally has exactly three occurrences of each 
variable: one in a clause of size three, and two in clauses of size two.
This $\phi'$ is then used to produce an input $\phi''$ to {\sc NaeSat*}.

\begin{enumerate}
\item Constructing $\phi'$.

Suppose variable $x_i$ appears $k \geq 2$ times in $\phi$. Create $k$ 
variables $x_{i1}, \ldots, x_{ik}$ for use in $\phi'$: use the same 
clauses, but replace each occurrence of $x_i$ by one of the $x_{ij}$. 
To enforce agreement between the different copies $x_{ij}$, add $k$ additional
clauses $(\overline{x}_{i1} \vee x_{i2}), (\overline{x}_{i2} \vee x_{i3}), 
\ldots, (\overline{x}_{ik}, x_{i1})$. These correspond to the implications
$x_1 \Rightarrow x_2, x_2 \Rightarrow x_3, \ldots, x_k \Rightarrow x_1$.

By design, $\phi$ is satisfiable if and only if $\phi'$ is satisfiable.

\item Constructing $\phi''$.

Now we construct an input $\phi''$ for {\sc NaeSat*}. Suppose $\phi'$ 
has $m$ clauses with three literals and $m'$ clauses with two literals. 
Create $2m+m'+1$ new variables: $s_1, \ldots, s_m$ and $f_1, \ldots, f_{m+m'}$
and $f$. 

If the $j$th three-literal clause in $\phi'$ is $(\alpha \vee \beta \vee \gamma)$,
replace it with two clauses in $\phi''$: $(\alpha \vee \beta \vee s_j)$ and
$(\overline{s}_j \vee \gamma \vee f_j)$. If the $j$th two-literal clause in
$\phi'$ is $(\alpha \vee \beta)$, replace it with $(\alpha \vee \beta \vee f_{m+j})$
in $\phi''$. Finally, add $m+m'$ clauses that enforce agreement among
the $f_i$: $(\overline{f}_1 \vee f_2 \vee f), (\overline{f}_2 \vee f_3 \vee f), 
\ldots, (\overline{f}_{m+m'} \vee f_1 \vee f)$.

All clauses in $\phi''$ have exactly three literals. Moreover, the only pairs 
of variables that occur together (in clauses) more than once are $\{f_i, f\}$ 
pairs. Each such pair occurs twice, as $\{f_i, f\}$ and $\{\overline{f}_i, f\}$. 
\end{enumerate}

\begin{lemma}
$\phi'$ is satisfiable if and only if $\phi''$ is not-all-equal satisfiable.
\end{lemma}

\begin{proof}
First suppose that $\phi'$ is satisfiable. Use the same settings of the
variables for $\phi''$. Set $f = f_1 = \cdots = f_{m+m'} = \mbox{\tt false}$.
For the $j$th three-literal clause $(\alpha \vee \beta \vee \gamma)$ of $\phi'$, 
if $\alpha = \beta = \mbox{\tt false}$ then set $s_j$ to {\tt true}, otherwise
set $s_j$ to {\tt false}. The resulting assignment satisfies exactly one
or two literals of each clause in $\phi''$.

Conversely, suppose $\phi''$ is not-all-equal satisfiable. Without loss
of generality, the satisfying assignment has $f$ set to {\tt false}
(otherwise flip all assignments). The clauses of the form 
$(\overline{f}_i \vee f_{i+1} \vee f)$ then enforce agreement among all
the $f_i$ variables. We can assume they are all {\tt false} (otherwise,
once again, flip all assignments). This means the two-literal clauses of 
$\phi'$ must be satisfied. Finally, consider any three-literal clause 
$(\alpha \vee \beta \vee \gamma)$ of $\phi'$. This was replaced by
$(\alpha \vee \beta \vee s_j)$ and $(\overline{s}_j \vee \gamma \vee f_j)$ 
in $\phi''$. Since $f_j$ is {\tt false}, it follows that one of the literals 
$\alpha, \beta, \gamma$ must be satisfied. Thus $\phi'$ is satisfied.
\end{proof}

\subsection{Hardness of {\sc Generalized 2-means}}

Given an instance $\phi(x_1, \ldots, x_n)$ of {\sc NaeSat*}, we construct a 
$2n \times 2n$ distance matrix $D = D(\phi)$ where the (implicit) $2n$ points 
correspond to literals. Entries of this matrix will be indexed as 
$D_{\alpha, \beta}$, for 
$\alpha, \beta \in \{x_1, \ldots, x_n, \overline{x}_1, \ldots, \overline{x}_n\}$. 
Another bit of notation: we write $\alpha \sim \beta$ to mean that either $\alpha$ 
and $\beta$ occur together in a clause or $\overline{\alpha}$ and $\overline{\beta}$
occur together in a clause. For instance, the clause $(x \vee \overline{y} \vee z)$
allows one to assert $\overline{x} \sim y$ but not $x \sim y$. The input restrictions
on {\sc NaeSat*} ensure that every relationship $\alpha \sim \beta$ 
is generated by a unique clause; it is not possible to have two different clauses
that both contain either $\{\alpha, \beta\}$ or 
$\{\overline{\alpha}, \overline{\beta}\}$.

Define
$$ D_{\alpha, \beta} = \left\{ 
\begin{array}{ll}
0          & \mbox{if $\alpha = \beta$} \\
1 + \Delta & \mbox{if $\alpha = \overline{\beta}$} \\
1 + \delta & \mbox{if $\alpha \sim \beta$} \\
1          & \mbox{otherwise}
\end{array}
\right.$$
Here $0 < \delta < \Delta < 1$ are constants such that 
$4 \delta m < \Delta \leq 1-2\delta n$, where $m$ is the 
number of clauses of $\phi$. One valid setting is $\delta = 1/(5m+2n)$ and 
$\Delta = 5\delta m$. 

\begin{lemma}
If $\phi$ is a satisfiable instance of {\sc NaeSat*}, then $D(\phi)$ admits a 
generalized 2-means clustering of cost $c(\phi) = n-1 + 2 \delta m/n$, where $m$ 
is the number of clauses of $\phi$.
\end{lemma}

\begin{proof}
The obvious clustering is to make one cluster (say $C_1$) consist of the
positive literals in the satisfying not-all-equal assignment and the other cluster 
($C_2$) the negative literals. 
Each cluster has $n$ points, and the distance between any two distinct points 
$\alpha, \beta$ within a cluster is either $1$ or, if $\alpha \sim \beta$,  $1 + \delta$.
Each clause of $\phi$ has at least one literal in $C_1$ and at least one literal
in $C_2$, since it is a not-all-equal assignment. Hence it contributes exactly one 
$\sim$ pair to $C_1$ and one $\sim$ pair to $C_2$.
The figure below shows an example with a clause $(x \vee \overline{y} \vee z)$ and
assignment $x = \mbox{\tt true}, y = z = \mbox{\tt false}$.

\begin{center}
\resizebox{2in}{!}{\begin{picture}(0,0)%
\includegraphics{clause.pstex}%
\end{picture}%
\setlength{\unitlength}{4144sp}%
\begingroup\makeatletter\ifx\SetFigFontNFSS\undefined%
\gdef\SetFigFontNFSS#1#2#3#4#5{%
  \reset@font\fontsize{#1}{#2pt}%
  \fontfamily{#3}\fontseries{#4}\fontshape{#5}%
  \selectfont}%
\fi\endgroup%
\begin{picture}(3540,4344)(391,-3988)
\put(406,-196){\makebox(0,0)[lb]{\smash{{\SetFigFontNFSS{14}{16.8}{\familydefault}{\mddefault}{\updefault}{\color[rgb]{0,0,0}$C_1$}%
}}}}
\put(3916,-151){\makebox(0,0)[lb]{\smash{{\SetFigFontNFSS{14}{16.8}{\familydefault}{\mddefault}{\updefault}{\color[rgb]{0,0,0}$C_2$}%
}}}}
\put(1179,-2753){\makebox(0,0)[lb]{\smash{{\SetFigFontNFSS{14}{16.8}{\familydefault}{\mddefault}{\updefault}{\color[rgb]{0,0,0}$\overline{z}$}%
}}}}
\put(2926,-2071){\makebox(0,0)[lb]{\smash{{\SetFigFontNFSS{14}{16.8}{\familydefault}{\mddefault}{\updefault}{\color[rgb]{0,0,0}$\overline{x}$}%
}}}}
\put(1171,-384){\makebox(0,0)[lb]{\smash{{\SetFigFontNFSS{14}{16.8}{\familydefault}{\mddefault}{\updefault}{\color[rgb]{0,0,0}$x$}%
}}}}
\put(1186,-1658){\makebox(0,0)[lb]{\smash{{\SetFigFontNFSS{14}{16.8}{\familydefault}{\mddefault}{\updefault}{\color[rgb]{0,0,0}$\overline{y}$}%
}}}}
\put(2941,-1029){\makebox(0,0)[lb]{\smash{{\SetFigFontNFSS{14}{16.8}{\familydefault}{\mddefault}{\updefault}{\color[rgb]{0,0,0}$z$}%
}}}}
\put(2955,-3368){\makebox(0,0)[lb]{\smash{{\SetFigFontNFSS{14}{16.8}{\familydefault}{\mddefault}{\updefault}{\color[rgb]{0,0,0}$y$}%
}}}}
\end{picture}%
}
\end{center}

Thus the clustering cost is
\begin{eqnarray*}
\frac{1}{2n} \sum_{i, i' \in C_1} D_{ii'} + \frac{1}{2n} \sum_{i, i' \in C_2} D_{ii'}
& = & 
 2 \cdot \frac{1}{n} \left( {n \choose 2} + m \delta \right) \\
& = & n-1 + \frac{2 \delta m}{n} .
\end{eqnarray*} 
\end{proof}

\begin{lemma}
Let $C_1, C_2$ be any 2-clustering of $D(\phi)$. If $C_1$ contains both
a variable and its negation, then the cost of this clustering is  
at least $n-1 + \Delta/(2n) > c(\phi)$.
\end{lemma} 

\begin{proof}
Suppose $C_1$ has $n'$ points while $C_2$ has $2n - n'$ points. Since all distances 
are at least $1$, and since $C_1$ contains a pair of points at distance $1 + \Delta$, 
the total clustering cost is at least
\begin{eqnarray*}
\frac{1}{n'} \left( {{n'} \choose 2} + \Delta \right) + \frac{1}{2n-n'} {{2n - n'} \choose 2} 
& = & n - 1 + \frac{\Delta}{n'} 
\ \geq \ n - 1 + \frac{\Delta}{2n}
 .
\end{eqnarray*}
Since $\Delta > 4 \delta m$, this is always more than $c(\phi)$. 
\end{proof}

\begin{lemma}
If $D(\phi)$ admits a 2-clustering of cost $\leq c(\phi)$, then $\phi$ is a 
satisfiable instance of {\sc NaeSat*}.
\end{lemma}

\begin{proof}
Let $C_1, C_2$ be a 2-clustering of cost $\leq c(\phi)$. By the previous lemma,
neither $C_1$ nor $C_2$ contain both a variable and its negation. Thus 
$|C_1| = |C_2| = n$. The cost of the clustering can be written as
$$ \frac{2}{n} \left({n \choose 2} + 
\delta\, \sum_{\mbox{clauses}} 
\left\{ \begin{array}{ll} 1 & \mbox{if clause split between $C_1$, $C_2$} \\
                          3 & \mbox{otherwise}
         \end{array} \right\} \right)$$ 
Since the cost is $\leq c(\phi)$, it follows that {\it all} clauses are split 
between $C_1$ and $C_2$, that is, every clause has at least one literal in 
$C_1$ and one literal in $C_2$. Therefore, the assignment that sets all of 
$C_1$ to {\tt true} and all of $C_2$ to {\tt false} is a valid {\sc NaeSat*} 
assignment for $\phi$.
\end{proof}

\subsection{Embeddability of $D(\phi)$}

We now show that $D(\phi)$ can be embedded into $l_2^2$, in the sense that there 
exist points $x_\alpha \in \R^{2n}$ such that $D_{\alpha,\beta} = \|x_\alpha - x_\beta\|^2$ for all $\alpha, \beta$. 
We rely upon the following classical result \cite{S38}.
\begin{thm}[Schoenberg]
Let $H$ denote the matrix $I - (1/N) {\bf 1}{\bf 1}^T$. An $N \times N$ symmetric matrix $D$ can be embedded into $l_2^2$ if and only if $-HDH$ is positive semidefinite.
\end{thm}
The following corollary is immediate.
\begin{cor}
An $N \times N$ symmetric matrix $D$ can be embedded into $l_2^2$ if and only if $u^T D u \leq 0$ for all $u \in \R^N$ with $u \cdot {\bf 1} = 0$.
\end{cor}
\begin{proof}
Since the range of the map $v \mapsto Hv$ is precisely $\{u \in \R^N: u \cdot {\bf 1} = 0\}$, we have
\begin{eqnarray*}
\mbox{$-HDH$ is positive semidefinite}
& \Leftrightarrow &
\mbox{$v^T HDH v \leq 0$ for all $v \in \R^N$} \\
& \Leftrightarrow &
\mbox{$u^T D u \leq 0$ for all $u \in \R^N$ with $u \cdot {\bf 1} = 0$} .
\end{eqnarray*}
\end{proof}

\begin{lemma}
$D(\phi)$ can be embedded into $l_2^2$.
\end{lemma}
\begin{proof}
If $\phi$ is a formula with variables $x_1, \ldots, x_n$, then $D = D(\phi)$ is a $2n \times 2n$ matrix whose first $n$ rows/columns correspond to $x_1, \ldots, x_n$ and remaining rows/columns correspond to $\overline{x}_1, \ldots, \overline{x}_n$. The entry for literals $(\alpha, \beta)$ is
$$ D_{\alpha \beta} 
= 1 
- {\bf 1}(\alpha = \beta) 
+ \Delta \cdot {\bf 1}(\alpha = \overline{\beta})
+ \delta \cdot {\bf 1}(\alpha \sim \beta),
$$
where ${\bf 1}(\cdot)$ denotes the indicator function.

Now, pick any $u \in \R^{2n}$ with $u \cdot {\bf 1} = 0$. Let $u^+$ denote the first $n$ coordinates of $u$ and $u^-$ the last $n$ coordinates.
\begin{eqnarray*}
u^T D u
& = & 
\sum_{\alpha, \beta} D_{\alpha \beta} u_\alpha u_\beta \\
& = & 
\sum_{\alpha, \beta} u_\alpha u_\beta 
\left( 1 - {\bf 1}(\alpha = \beta) + \Delta \cdot {\bf 1}(\alpha = \overline{\beta})
+ \delta \cdot {\bf 1}(\alpha \sim \beta) \right) \\
& = & 
\sum_{\alpha, \beta} u_\alpha u_\beta
- \sum_{\alpha} u_\alpha^2 
+ \Delta \sum_{\alpha} u_\alpha u_{\overline{\alpha}}
+ \delta \sum_{\alpha, \beta} u_\alpha u_\beta {\bf 1}(\alpha \sim \beta) \\
& \leq & 
\left( \sum_\alpha u_\alpha \right)^2 - \|u\|^2 + 2 \Delta (u^+ \cdot u^-) 
+ \delta \sum_{\alpha, \beta} |u_\alpha| |u_\beta| \\
& \leq &
- \|u\|^2 + \Delta (\|u^+\|^2 + \|u^-\|^2) 
+ \delta \left( \sum_{\alpha} |u_\alpha| \right)^2 \\
& \leq & 
- (1 - \Delta) \|u\|^2 + 2 \delta \|u\|^2 n
\end{eqnarray*}
where the last step uses the Cauchy-Schwarz inequality. Since
$2 \delta n \leq 1-\Delta$, this quantity is always $\leq 0$.
\end{proof}

\end{document}